\documentclass[runningheads]{llncs}

\usepackage{amsmath}  %
\usepackage[T1]{fontenc}
\usepackage{newtxtext}
\usepackage[varvw]{newtxmath}
\usepackage{graphicx}
\usepackage{booktabs}
\usepackage{multirow}
\usepackage{makecell}
\usepackage{xcolor}
\usepackage{xspace}
\usepackage{tikz}
\usetikzlibrary{positioning,arrows.meta,calc}
\usepackage[hidelinks]{hyperref}

\newcommand{\wsms}{whole-slice misses\xspace}

\newcommand{\lost}{lost slice\xspace}
\newcommand{\losts}{lost slices\xspace}

\newcommand{\offgate}{official gate\xspace}
\newcommand{\symgate}{symmetric gate\xspace}
\newcommand{\cnf}{Chain-and-Fill\xspace}         %
\newcommand{\chainonly}{chain-only\xspace}       %
\newcommand{\Chainonly}{Chain-only\xspace}
\newcommand{\nsi}{gated nearest-slice interpolation\xspace} %
\newcommand{\mcohort}{measurement cohort\xspace}    %
\newcommand{\ccohort}{confirmation cohort\xspace}   %

\newcommand{\Pone}{\textbf{P1}\xspace}  %
\newcommand{\Ptwo}{\textbf{P2}\xspace}  %

\newcommand{\areaThr}{256\xspace}      %
\newcommand{\lostThr}{10\%\xspace}     %
\newcommand{\recThr}{50\%\xspace}      %
\newcommand{\sevThr}{30\%\xspace}      %
\newcommand{\gateMean}{0.85\xspace}    %
\newcommand{\gateOrgan}{0.7\xspace}    %

\definecolor{ink}{HTML}{1A1A1A}
\definecolor{okgrey}{HTML}{6E6E6E}
\definecolor{okblue}{HTML}{0072B2}
\definecolor{okgreen}{HTML}{009E73}
\definecolor{oklost}{HTML}{D55E00}

\newsavebox{\figoneleft}

\newcommand{\pci}[2]{\makecell{#1\\#2}}

\newcommand{\amosN}{120\xspace} %
\newcommand{\amosSevereN}{36\xspace} %
\newcommand{\amosSeverePct}{30.0\%\xspace} %
\newcommand{\amosSevereCI}{22.0--39.0\%\xspace} %
\newcommand{\amosLostVoxShare}{71.8\%\xspace} %
\newcommand{\amosSevereLostShareMed}{82.9\%\xspace} %
\newcommand{\amosSpleenMidDicePct}{1.64\%\xspace} %
\newcommand{\amosSpleenVolMinusSlice}{0.108\xspace} %
\newcommand{\amosOrgansMidDiceMin}{1.0\%\xspace} %
\newcommand{\amosOrgansMidDiceMax}{2.6\%\xspace} %
\newcommand{\amosSpleenLowDiceN}{767\xspace} %
\newcommand{\amosSpleenLowDicePct}{25.1\%\xspace} %
\newcommand{\rtLostN}{773\xspace} %
\newcommand{\rtControlN}{2{,}269\xspace} %
\newcommand{\rtOfficialLostPass}{83.31\%\xspace} %
\newcommand{\rtOfficialLostPassCI}{78.89--87.50\%\xspace} %
\newcommand{\rtSymLostPass}{68.82\%\xspace} %
\newcommand{\rtSymLostPassCI}{64.43--73.69\%\xspace} %
\newcommand{\rtOfficialControlPass}{91.58\%\xspace} %
\newcommand{\rtSymControlPass}{89.73\%\xspace} %
\newcommand{\rtBlindReproduced}{88.03\%\xspace} %
\newcommand{\rtBlindReproducedCI}{84.31--91.21\%\xspace} %
\newcommand{\rtBlindN}{702\xspace} %
\newcommand{\rtReproducedN}{618\xspace} %
\newcommand{\rtOfficialGap}{8.3\xspace} %
\newcommand{\rtOfficialGapCI}{4.1--12.5\xspace} %
\newcommand{\rtSymGap}{20.9\xspace} %
\newcommand{\rtSymGapCI}{16.2--25.3\xspace} %
\newcommand{\rtOfficialLostFlag}{16.7\%\xspace} %
\newcommand{\rtSymLostFlag}{31.2\%\xspace} %
\newcommand{\rtOfficialControlFlag}{8.4\%\xspace} %
\newcommand{\rtSymControlFlag}{10.3\%\xspace} %
\newcommand{\rtBlindShare}{90.8\%\xspace} %
\newcommand{\rtBlindOfficialPass}{90.9\%\xspace} %
\newcommand{\rtScoredN}{71\xspace} %
\newcommand{\rtScoredPassN}{6\xspace} %
\newcommand{\rtScoredPassPct}{8.5\%\xspace} %
\newcommand{\rtVisibleN}{155\xspace} %
\newcommand{\rtVisiblePct}{20.1\%\xspace} %
\newcommand{\zInternalLostFlag}{42.2\%\xspace} %
\newcommand{\zInternalMissedOutside}{95.75\%\xspace} %
\newcommand{\noisePersist}{85.5\%\xspace} %
\newcommand{\noiseRecoveredN}{90\xspace} %
\newcommand{\noiseNewN}{110\xspace} %
\newcommand{\zInternalLostShare}{42.17\%\xspace} %
\newcommand{\gstSevereN}{41\xspace} %
\newcommand{\gstRecoveredN}{24\xspace} %
\newcommand{\gstStillLostN}{746\xspace} %
\newcommand{\gstNewLostN}{195\xspace} %
\newcommand{\gstLostTotalN}{941\xspace} %
\newcommand{\gstSpleenDiff}{$-$0.015\xspace} %
\newcommand{\gstSpleenDiffCI}{[$-$0.027, $-$0.003]\xspace} %
\newcommand{\gstSevereTableP}{0.18\xspace} %
\newcommand{\gstPersist}{96.5\%\xspace} %
\newcommand{\noisePartialN}{22\xspace} %
\newcommand{\gstPartialN}{3\xspace} %
\newcommand{\noiseStillN}{661\xspace} %
\newcommand{\gstRecFewer}{0.55\xspace} %
\newcommand{\gstRecFewerCI}{0.33--0.78\xspace} %
\newcommand{\gstNewMore}{0.71\xspace} %
\newcommand{\gstNewMoreCI}{0.47--0.95\xspace} %
\newcommand{\gstLostIncreasePct}{22\%\xspace} %
\newcommand{\organsMeanDiff}{+0.009\xspace} %
\newcommand{\pairLostN}{771\xspace} %
\newcommand{\pairStillN}{745\xspace} %
\newcommand{\pairRecoveredN}{19\xspace} %
\newcommand{\pairNewN}{224\xspace} %
\newcommand{\pairPersist}{96.6\%\xspace} %
\newcommand{\pairPersistCI}{94.7--98.3\%\xspace} %
\newcommand{\stageBN}{119\xspace} %
\newcommand{\lfourRecoveredN}{678\xspace} %
\newcommand{\lfourLostN}{1{,}784\xspace} %
\newcommand{\lfourRecoveredPct}{38.0\%\xspace} %
\newcommand{\lfourVsLone}{+0.040\xspace} %
\newcommand{\lfourVsLoneCI}{[+0.021, +0.066]\xspace} %
\newcommand{\cnfCone}{+0.116\xspace} %
\newcommand{\cnfConeCI}{[+0.078, +0.153]\xspace} %
\newcommand{\cnfCtwo}{+0.010\xspace} %
\newcommand{\cnfCthree}{0.702\xspace} %
\newcommand{\cnfCthreeCI}{[0.635, 0.785]\xspace} %
\newcommand{\cnfCfour}{+0.046\xspace} %
\newcommand{\cnfCfourCI}{[+0.038, +0.056]\xspace} %
\newcommand{\cnfHarmed}{1/357\xspace} %
\newcommand{\cnfMet}{4/4\xspace} %
\newcommand{\nsiCone}{+0.040\xspace} %
\newcommand{\nsiConeCI}{[+0.013, +0.076]\xspace} %
\newcommand{\nsiCtwo}{+0.003\xspace} %
\newcommand{\nsiCthree}{0.616\xspace} %
\newcommand{\nsiCthreeCI}{[0.523, 0.710]\xspace} %
\newcommand{\nsiCfour}{+0.018\xspace} %
\newcommand{\nsiCfourCI}{[+0.012, +0.025]\xspace} %
\newcommand{\nsiHarmed}{7/357\xspace} %
\newcommand{\nsiMet}{3/4\xspace} %
\newcommand{\chainonlyCone}{+0.003\xspace} %
\newcommand{\chainonlyConeCI}{[+0.001, +0.003]\xspace} %
\newcommand{\chainonlyCtwo}{+0.003\xspace} %
\newcommand{\chainonlyCfour}{+0.005\xspace} %
\newcommand{\chainonlyCfourCI}{[+0.001, +0.008]\xspace} %
\newcommand{\chainonlyHarmed}{8/357\xspace} %
\newcommand{\chainonlyMet}{2/4\xspace} %
\newcommand{\seedZeroPersist}{97.0\%\xspace} %
\newcommand{\seedZeroSpleenDiff}{$-$0.012\xspace} %
\newcommand{\seedOnePersist}{91.1\%\xspace} %
\newcommand{\seedOneSpleenDiff}{$-$0.014\xspace} %
\newcommand{\seedTwoPersist}{88.3\%\xspace} %
\newcommand{\seedTwoSpleenDiff}{$-$0.003\xspace} %
\newcommand{\reproSource}{85.04\%\xspace} %
\newcommand{\reproTrained}{86.93\%\xspace} %
\newcommand{\reproPaperTrained}{87.2\%\xspace} %
\newcommand{\reproPaperSource}{84.4\%\xspace} %
\newcommand{\caseLostN}{12\xspace} %
\newcommand{\caseEligN}{23\xspace} %
\newcommand{\caseLostPassN}{8\xspace} %
\newcommand{\caseDice}{0.51\xspace} %
\newcommand{\areaMmRange}{76--274\xspace} %
\newcommand{\samtwoCone}{+0.105\xspace} %
\newcommand{\samtwoConeCI}{[+0.047, +0.161]\xspace} %
\newcommand{\samtwoCtwo}{+0.002\xspace} %
\newcommand{\samtwoCthree}{0.383\xspace} %
\newcommand{\samtwoCthreeCI}{[0.327, 0.448]\xspace} %
\newcommand{\samtwoCfour}{+0.014\xspace} %
\newcommand{\samtwoCfourCI}{[$-$0.004, +0.033]\xspace} %
\newcommand{\samtwoHarmed}{63/357\xspace} %
\newcommand{\samtwoMet}{2/4\xspace} %
\newcommand{\interpCone}{+0.013\xspace} %
\newcommand{\interpConeCI}{[+0.003, +0.051]\xspace} %
\newcommand{\interpCtwo}{+0.004\xspace} %
\newcommand{\interpCthree}{0.860\xspace} %
\newcommand{\interpCthreeCI}{[0.791, 0.926]\xspace} %
\newcommand{\interpCfour}{+0.018\xspace} %
\newcommand{\interpCfourCI}{[+0.012, +0.023]\xspace} %
\newcommand{\interpHarmed}{8/357\xspace} %
\newcommand{\interpMet}{3/4\xspace} %
\newcommand{\unionCone}{+0.063\xspace} %
\newcommand{\unionConeCI}{[+0.045, +0.096]\xspace} %
\newcommand{\unionCtwo}{$-$0.022\xspace} %
\newcommand{\unionCthree}{0.412\xspace} %
\newcommand{\unionCthreeCI}{[0.371, 0.453]\xspace} %
\newcommand{\unionCfour}{+0.007\xspace} %
\newcommand{\unionCfourCI}{[$-$0.001, +0.015]\xspace} %
\newcommand{\unionHarmed}{23/357\xspace} %
\newcommand{\unionMet}{1/4\xspace} %
\newcommand{\dSamTwo}{0.000\xspace} %
\newcommand{\dSamTwoCI}{[$-$0.033, +0.048]\xspace} %
\newcommand{\dInterp}{+0.059\xspace} %
\newcommand{\dInterpCI}{[+0.036, +0.105]\xspace} %
\newcommand{\dUnion}{+0.032\xspace} %
\newcommand{\dUnionCI}{[+0.007, +0.054]\xspace} %
\newcommand{\cnfRecInner}{70\%\xspace} %
\newcommand{\cnfRecOuter}{29\%\xspace} %
\newcommand{\samtwoRecInner}{96\%\xspace} %
\newcommand{\samtwoRecOuter}{92\%\xspace} %
\newcommand{\ccSourceSevereN}{25\xspace} %
\newcommand{\ccSourceSevereCI}{14.1--29.4\%\xspace} %
\newcommand{\ccSourceLostN}{462\xspace} %
\newcommand{\ccSeedLostRange}{540--633\xspace} %
\newcommand{\mtwoSpleenDice}{0.007\xspace} %
\newcommand{\mtwoGateDice}{0.91\xspace} %
\newcommand{\dfgSpleenDice}{0.48\xspace} %
\newcommand{\dfgFourDice}{0.62\xspace} %
\newcommand{\pcSpleenDice}{0.51\xspace} %
\newcommand{\pcFourDice}{0.58\xspace} %
\newcommand{\targetGateDice}{0.70\xspace} %
\newcommand{\nnqcSliceAUROC}{0.31\xspace} %
\newcommand{\nnqcSliceAUROCCI}{0.27--0.34\xspace} %
\newcommand{\nnqcScoredAUROC}{0.34\xspace} %
\newcommand{\nnqcPatientAUROC}{0.66\xspace} %
\newcommand{\nnqcPatientAUROCCI}{0.56--0.76\xspace} %
\newcommand{\nnqcLostBelow}{89.4\%\xspace} %
\newcommand{\nnqcLostN}{611\xspace} %
\newcommand{\btcvN}{24\xspace} %
\newcommand{\btcvSevereN}{4\xspace} %
\newcommand{\btcvLostN}{92\xspace} %
\newcommand{\btcvLostPassN}{79\xspace} %
\newcommand{\btcvLostPass}{85.9\%\xspace} %
\newcommand{\btcvControlPass}{93.2\%\xspace} %
\newcommand{\btcvBlindN}{85\xspace} %
\newcommand{\btcvReproducedN}{72\xspace} %
\newcommand{\btcvReproduced}{84.7\%\xspace} %

\newif\ifanon
\anonfalse

\begin{document}

\title{Whole-Slice Misses Escape Round-Trip Filtering\texorpdfstring{\\}{ } and Volumetric Dice in a Source-Free\texorpdfstring{\\}{ } Flow-Matching Segmenter}
\titlerunning{Whole-Slice Misses}

\ifanon
  \author{Anonymized Authors}
  \authorrunning{Anonymized Authors}
  \institute{Anonymized Affiliations}
\else
  \author{Guobiao Wang}
  \authorrunning{G. Wang}
  \institute{Columbia University, New York, NY, USA\\ \email{gw2512@columbia.edu}}
\fi

\maketitle

\begin{abstract}
SymmAdapt, a source-free flow-matching segmentation pipeline, relies on two quality gates: a
round-trip filter on pseudo-labels and volumetric Dice in evaluation. We audit both on \amosN
AMOS CT patients unseen in training and adaptation. The model often misses the spleen on whole
slices, severely so in \amosSeverePct of patients. The round-trip filter passes
\rtOfficialLostPass of these lost slices, because its round trip reproduces the miss and the
filter scores only organs present in the pseudo-label. Volumetric Dice depends on how much volume
is missed, not where, and self-training leaves \gstPersist of lost slices lost (\pairPersist on
a second noise stream). A model-free repair, \cnf, met all pre-registered endpoints on
\stageBN further patients and could not be distinguished from SAM\,2 propagation in
severe-group gain, with \cnfHarmed rather than \samtwoHarmed harmed patient--seed pairs. Reporting lost slices next to volumetric Dice makes the error visible at
no cost.
\end{abstract}

\section{Introduction}
\label{sec:intro}

Source-free adaptation lets a segmenter trained on one modality be adapted to another
without the source data, typically by self-training on its own pseudo-labels. SymmAdapt
does this for the slice-wise flow-matching segmenter SymmFlow, adapting from MRI to
CT~\cite{symmadapt2026,caetano2026symmflow}. Two quality gates judge the result. A round-trip
filter keeps a pseudo-label slice only if an image rendered from it is segmented back into
the same mask, and volumetric Dice summarises each patient in the evaluation.

\begin{figure}[t]
  \centering
  \sbox{\figoneleft}{\begin{tikzpicture}[font=\scriptsize, x=1mm, y=1mm]
  \def\h{2.6}   %
  \def\g{0.5}   %
  \newcommand{\stack}[3]{%
    \foreach \w [count=\k] in {4,6.4,8,8,6.4,4} {
      \pgfmathsetmacro{\yy}{-(\k-1)*(\h+\g)}
      \ifnum\k>4 \ifnum#3=1
        \draw[oklost, line width=0.6pt, dash pattern=on 1.2pt off 0.8pt, rounded corners=0.6pt]
          ({#1-\w/2},\yy) rectangle ({#1+\w/2},\yy-\h);
      \else
        \draw[#2, rounded corners=0.6pt] ({#1-\w/2},\yy) rectangle ({#1+\w/2},\yy-\h);
      \fi \else
        \draw[#2, rounded corners=0.6pt] ({#1-\w/2},\yy) rectangle ({#1+\w/2},\yy-\h);
      \fi
    }}
  \fill[okgrey!14, rounded corners=1.2pt] (6.8,0.4) rectangle (35.6,-19.3);
  \node[anchor=north, inner sep=1pt, text=ink!75] at (21.2,-19.4) {round-trip gate};
  \stack{-2.5}{draw=ink, line width=0.6pt}{0}
  \stack{12}{draw=ink, line width=0.4pt, fill=okgrey!45}{1}
  \stack{29}{draw=ink, line width=0.4pt, fill=okgrey!45}{1}
  \stack{56}{draw=ink, line width=0.4pt, fill=okgrey!45}{1}
  \node[above] at (-2.5,0.6)  {Reference};
  \node[above] at (12,0.6) {Pseudo-label $y$};
  \node[above] at (29,0.6) {Round trip $\hat y$};
  \node[above] at (56,0.6) {Self-trained};
  \draw[-{Stealth[length=1.4mm]}, line width=0.5pt] (16.6,-7.8) -- (24.4,-7.8);
  \draw[-{Stealth[length=1.6mm]}, line width=0.5pt] (37.0,-7.8) -- (51.4,-7.8);
  \node[above, align=center] at (44.2,-7.4) {pass: $y$\\[-1pt] is the target};
  \node[below, align=center] at (44.2,-8.2) {spleen =\\[-1pt] background};
  \draw[line width=0.5pt] (32.9,-12.4) -- (33.5,-12.4) -- (33.5,-18.1) -- (32.9,-18.1);
  \draw[line width=0.5pt] (59.9,-12.4) -- (60.5,-12.4) -- (60.5,-18.1) -- (59.9,-18.1);
  \node[anchor=west, align=left, inner sep=1pt] at (60.8,-15.2) {\gstPersist\\[-1pt] stay lost\\[-1pt] (AMOS)};
  \node[anchor=north west, align=left, inner sep=0pt] at (-7,-23.4)
    {vol.\ Dice 0.84,\\[-1pt] slice mean 0.67 \textbf{(P2)}};
  \node[anchor=north, align=center, inner sep=0pt] at (29,-23.4)
    {absent in $y$ and $\hat y$:\\[-1pt] not scored \textbf{(P1)}};
  \node[anchor=north, align=center, inner sep=0pt] at (54,-23.4)
    {BTCV: \btcvLostPassN{} of \btcvLostN{} lost\\[-1pt] slices became targets};
\end{tikzpicture}}%
  \usebox{\figoneleft}\hspace{4mm}%
  \includegraphics[height=\dimexpr\ht\figoneleft+\dp\figoneleft\relax]{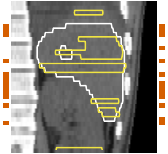}
  \caption{How a lost slice escapes the pipeline.
    Left: the round trip reproduces the miss, so the gate never scores the organ (\Pone).
    Self-training then learns the miss as background, and volumetric Dice exceeds the slice
    mean (\Ptwo).
    Right: the severe patient with the median number of lost slices on the \mcohort, chosen
    by this rule before viewing. White: reference; yellow: source model. Ticks mark lost slices
    of the source model (left; \caseLostN among \caseEligN, \caseLostPassN pass the \offgate,
    volumetric Dice \caseDice) and of the self-trained model (right).}
  \label{fig:overview}
\end{figure}

Both gates miss one kind of error. On unseen patients, the model leaves out the spleen on
whole axial slices while segmenting the neighbouring slices well (Fig.~\ref{fig:overview}).
We call these \wsms. In \amosSevereN of \amosN patients the model misses at least \sevThr of
the spleen, with a median \amosSevereLostShareMed of the missed volume on such slices. Any use
that reads the mask slice by slice, such as organ volume measured over a slice range, inherits
them. Yet volumetric Dice records them only as missing volume, and the round-trip filter passes
most of the affected slices.

A 2D network is known to miss structures on difficult slices. Zheng et al.\ documented this in
cardiac MRI and measured it with a presence rate~\cite{zheng2018consistent}, and SDC-UDA
plotted slice Dice to show sudden slice-to-slice fluctuations in earlier cross-modality
adaptation methods~\cite{shin2023sdcuda}. Agreement is also known not to establish correctness.
Ensemble members can agree on a wrongly segmented region~\cite{zenk2025failure}, confidence
scores do not reliably catch silent failures~\cite{bungert2023silent}, and
distribution-matching translation can add or remove image features~\cite{cohen2018hallucinate}.
Aggregated metrics, in turn, can hide errors on small
structures~\cite{reinke2024pitfalls,maierhein2024metrics}. Neither line of work shows why the
two gates of this pipeline cannot see \wsms, or what self-training does to them. We answer both
questions with pre-registered analyses on two disjoint cohorts, the second opened only after the
repair was frozen.

We make four contributions:
\begin{enumerate}
  \item We quantify \wsms on \amosN unseen patients: \amosSeverePct are severe, and slice
    Dice is all-or-none (Sec.~\ref{sec:phenomenon}).
  \item We show that the round trip reproduces the miss in \rtBlindReproduced of lost slices
    without spleen, which hides it from any gate that scores agreement between a pseudo-label
    and its round trip, whatever the threshold. The official gate adds a second blind spot by
    scoring only organs present in the pseudo-label (\Pone), which scoring organs present in
    either mask removes (Sec.~\ref{sec:roundtrip}).
  \item We show that self-training leaves \gstPersist of lost slices lost (\pairPersist on a
    second noise stream), while volumetric Dice, which depends on how much volume is missed and
    not where (\Ptwo), barely moves (Secs.~\ref{sec:phenomenon},~\ref{sec:selftraining}).
  \item We show that a model-free repair reduces the misses. \cnf, confirmed on \stageBN
    disjoint patients and three seeds, could not be distinguished from SAM\,2 propagation in
    severe-group gain, harmed fewer patient--seed pairs, and exceeded interpolation and
    multi-sample baselines (Sec.~\ref{sec:repair}).
\end{enumerate}

\section{Setting and Definitions}
\label{sec:setting}

\paragraph{Pipeline.}
We study SymmAdapt~\cite{symmadapt2026}, a source-free adaptation of the slice-wise
flow-matching segmenter SymmFlow~\cite{caetano2026symmflow}, in its released abdominal
MRI$\to$CT setting. A source model trained on CHAOS MRI~\cite{kavur2021chaos} is self-trained
on unlabelled BTCV CT~\cite{landman2015btcv} with pseudo-label slices that pass a round-trip
filter (Sec.~\ref{sec:roundtrip}). The filter requires a mean Dice over scored organs of at
least \gateMean and at least \gateOrgan for every scored organ, with the optional
image-similarity checks disabled. We use
the released source weights, reproduce self-training with the official code and settings on
data preprocessed as in DFG~\cite{huai2025dfg}, and compute all metrics on the
$512\times512$ inference grid. On the original six BTCV test patients, our source and
self-trained models reach a four-organ mean Dice of \reproSource and \reproTrained, against the
reported \reproPaperSource and \reproPaperTrained~\cite{symmadapt2026}.

\paragraph{Cohorts.}
All analyses use AMOS CT scans~\cite{ji2022amos}, unseen in training or adaptation. One random
permutation of the 300 labelled scans was split in order. The first five patients checked the
preprocessing on the liver alone. The next \amosN form the \mcohort, used to measure the
phenomenon and to select the repair variant. The \ccohort of the next \stageBN was opened only
after the repair was frozen, and the last 55 form a reserve that was never read. One measurement
patient whose reference spleen never reached \areaThr pixels was excluded and replaced in
permutation order, which consumed one confirmation position. The round-trip filter acts on BTCV
during adaptation, and we audit it there on the \btcvN training patients
(Sec.~\ref{sec:roundtrip}). All other rates are measured on AMOS, to which no model was adapted.

\paragraph{Definitions.}
We analyse spleen slices whose reference area is at least \areaThr pixels
(\areaMmRange\,mm$^2$ on the inference grid). Such a slice is \emph{lost} when the prediction covers less than
\lostThr of the reference slice, and \emph{recovered} after an intervention when coverage
reaches \recThr. A patient is \emph{severe} when the model misses at least \sevThr of the
reference spleen voxels. Because severity is defined by the reference label, it serves only for
evaluation, and every severe-group result is reported with its whole-cohort counterpart. The
\offgate scores an organ only if it is present in the pseudo-label it checks, and the \symgate
scores it if it is present in the pseudo-label or its round trip (Sec.~\ref{sec:roundtrip}).

\paragraph{Statistics and registration.}
Proportions carry Clopper--Pearson 95\% intervals, and other intervals come from 2{,}000
patient-level bootstrap resamples stratified by severity. Each analysis was pre-registered with
its thresholds and wording rules before its results were computed. Rules fixed after inspecting
development data are marked where they are used. Endpoints and comparisons are interpreted
separately, without multiplicity correction. Code, pre-registrations with their commit history,
and frozen summaries are available at \url{https://anonymous.4open.science/r/wsm-audit-8281}.

\paragraph{Data use.}
All datasets are public and de-identified. AMOS~\cite{ji2022amos} is released under CC BY 4.0.
CHAOS~\cite{kavur2021chaos} (CC BY-NC-SA 4.0) enters through the source weights and trains the
second models. BTCV~\cite{landman2015btcv}, acquired under institutional review board
supervision, is distributed through Synapse; we used the preprocessed form released with
DFG~\cite{huai2025dfg}.
No further ethics approval was required for this retrospective use of public data.

\section{Whole-Slice Misses and Why Volumetric Dice Under-Reports Them}
\label{sec:phenomenon}

\begin{figure}[t]
  \centering
  \includegraphics[width=\textwidth]{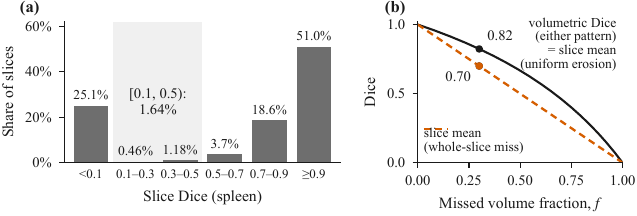}
  \caption{Slice Dice is all-or-none, and volumetric Dice cannot tell where volume is missed.
    (a)~Spleen slice Dice of the source model on the \mcohort (\amosN patients,
    slices with reference area ${\geq}256$ pixels); the shaded band is $[0.1,0.5)$.
    (b)~The same fraction $f$ of volume missed as whole slices or as uniform erosion of every
    slice (analytic, equal-area slices, no over-segmentation). Volumetric Dice is
    $2(1-f)/(2-f)$ for both patterns, and only the slice mean separates them.}
  \label{fig:bimodal}
\end{figure}

On \amosN unseen AMOS patients, \amosSevereN (\amosSeverePct; 95\% CI \amosSevereCI)
were severe, and \amosLostVoxShare of all missed spleen voxels lay on \losts.
Slice Dice was bimodal: \amosSpleenLowDicePct of spleen slices scored below 0.1,
but only \amosSpleenMidDicePct fell in $[0.1,0.5)$
(\amosOrgansMidDiceMin--\amosOrgansMidDiceMax across four organs; Fig.~\ref{fig:bimodal}a).
The coverage-based definition of a \lost (Sec.~\ref{sec:setting}) counts \rtLostN slices,
close to the \amosSpleenLowDiceN with Dice below 0.1, so the count does not hinge on where
the threshold is placed within the sparsely populated middle range.
The pattern recurred on the \ccohort, where the source model was severe in \ccSourceSevereN of
\stageBN patients (95\% CI \ccSourceSevereCI), with \ccSourceLostN lost slices.

\paragraph{Why volumetric Dice under-reports the error (\Ptwo).}
Let $P_i$ and $G_i$ be the predicted and reference organ in slice $i$, and
$d_i = 2|P_i\cap G_i|/(|P_i|+|G_i|)$ the slice Dice. Volumetric Dice is an
area-weighted mean of slice Dice,
\begin{equation}
  D_{\mathrm{vol}}
  = \frac{2\sum_i |P_i\cap G_i|}{\sum_i \bigl(|P_i|+|G_i|\bigr)}
  = \sum_i w_i\, d_i,
  \qquad
  w_i = \frac{|P_i|+|G_i|}{\sum_j \bigl(|P_j|+|G_j|\bigr)},
  \label{eq:dvol}
\end{equation}
where slices with $P_i = G_i = \emptyset$ have $w_i=0$ and drop out.
Without over-segmentation ($P_i\subseteq G_i$), Eq.~\eqref{eq:dvol} reduces to
$D_{\mathrm{vol}} = 2R/(1+R)$, where $R$ is the volumetric recall. The score depends on
how much volume is missed, not on where. Missing a fraction $f$ of the volume as whole
slices or as uniform erosion of every slice yields the same $D_{\mathrm{vol}}$,
whereas the slice mean separates the two patterns (Fig.~\ref{fig:bimodal}b).
On the \mcohort, mean spleen volumetric Dice exceeded the slice mean by
\amosSpleenVolMinusSlice.

\section{Auditing the Round-Trip Filter}
\label{sec:roundtrip}

\paragraph{What each gate can see (\Pone).}
The round-trip filter renders an image from the pseudo-label $y$, re-segments it to
obtain $\hat y$, and scores agreement per organ~\cite{symmadapt2026}. Dice is undefined
when the reference is empty, and the official implementation skips organs absent from
the pseudo-label being checked. With $\mathcal{K}$ the set of organs,
\begin{equation}
  \mathcal{K}_{\mathrm{off}} = \{k\in\mathcal{K} : k \in y\},
  \qquad
  \mathcal{K}_{\mathrm{sym}} = \{k\in\mathcal{K} : k \in y \cup \hat y\}
  \label{eq:gates}
\end{equation}
are the organs scored by the \offgate and by a \symgate that scores an organ present in
either mask. An organ missing from $y$ is never scored by the \offgate, so the slice passes
or fails on the remaining organs. The \symgate closes this gap only when the round trip
restores the organ. If $\hat y$ reproduces the miss, the organ is absent from both masks,
and no score of agreement between $y$ and $\hat y$ has anything to flag.
A reproduced miss is therefore invisible to every agreement-based gate, whatever its
threshold. The wording rules for this
analysis were registered after the official pass rate on lost slices was known, but before
any symmetric-gate or re-run result was computed.

\paragraph{Both mechanisms are present.}
On the \mcohort, \rtLostN spleen slices were lost. In \rtBlindN of them (\rtBlindShare) the
pseudo-label contained no spleen, so the \offgate could not score the organ, and these slices
passed at \rtBlindOfficialPass, as often as control slices (\rtOfficialControlPass). In the
\rtScoredN lost slices that kept a spleen fragment, the gate scored the organ and passed only
\rtScoredPassN (\rtScoredPassPct). The \offgate thus rejects the lost slices it can see, but
it sees fewer than one in ten. Overall it passed \rtOfficialLostPass (95\% CI
\rtOfficialLostPassCI) of lost slices against \rtOfficialControlPass of controls, a gap of
\rtOfficialGap percentage points (\rtOfficialGapCI).

The \symgate lowered the pass rate on lost slices to \rtSymLostPass (\rtSymLostPassCI) while
controls stayed at \rtSymControlPass, which widened the gap to \rtSymGap points (\rtSymGapCI).
It could not go much further, because in \rtReproducedN of the \rtBlindN slices without
spleen (\rtBlindReproduced, \rtBlindReproducedCI) the round trip reproduced the miss. A gate
that scores agreement between $y$ and $\hat y$ can reject a lost slice \emph{for the missing
spleen} only if the slice kept a fragment or its round trip restored the organ. This holds for
\rtVisibleN of \rtLostN lost slices (\rtVisiblePct). Every other rejection of a lost slice comes from
the remaining organs. The filter thus fails for two separable reasons: a reference blind spot
(\Pone), which the \symgate removes, and self-reproduction, which no agreement-based gate can
remove.

\paragraph{The same holds during adaptation.}
On the \btcvN BTCV patients used for self-training (\btcvSevereN severe), the \offgate passed
\btcvLostPassN of \btcvLostN lost slices (\btcvLostPass), all without spleen in the
pseudo-label, against \btcvControlPass of controls. Per-patient pass rates had been seen in
exploration before this audit was registered. A re-run of the stochastic round trip reproduced
the miss in \btcvReproducedN of the \btcvBlindN slices without spleen (\btcvReproduced).

\paragraph{A mask-only check covers the interior, not the ends.}
Flagging every empty slice between the first and last slices with a spleen component of at
least 256 pixels catches \zInternalLostShare of lost slices. Of the rest, \zInternalMissedOutside
lie beyond those slices, where the rule has no anchor on one side (Table~\ref{tab:gates}).

\paragraph{An image-based check does not single them out either.}
nnQC~\cite{marciano2025nnqc} generates a pseudo ground truth from the real image and scores a
mask by its Dice against it. Its released weights were trained with CT labels, which the
source-free setting does not allow. On the first self-trained seed on the \ccohort, nnQC scored
\nnqcLostBelow of the \nnqcLostN lost slices below 0.70, but control slices tended to score
lower still. The AUROC for separating lost from control slices was \nnqcSliceAUROC (95\% CI
\nnqcSliceAUROCCI), below chance, and \nnqcScoredAUROC without the slices nnQC left unscored. At
patient level, nnQC separated severe from non-severe patients with AUROC \nnqcPatientAUROC
(\nnqcPatientAUROCCI).

\begin{table}[t]
  \centering
  \caption{What each check can see. Rejection rates on the \mcohort
    (\rtLostN lost, \rtControlN control slices). None of the three looks at the real image.}
  \label{tab:gates}
  \small
  \setlength{\tabcolsep}{4pt}
  \begin{tabular}{@{}lccrr@{}}
    \toprule
    Check & \makecell{Organ absent\\from $y$} &
      \makecell{Reproduced\\miss} & \makecell{Lost\\rejected} & \makecell{Control\\rejected} \\
    \midrule
    Official round-trip gate  & no                   & no              & \rtOfficialLostFlag & \rtOfficialControlFlag \\
    Symmetric round-trip gate & if $\hat y$ restores & no              & \rtSymLostFlag      & \rtSymControlFlag \\
    Interior-gap rule (mask)  & between anchors      & between anchors & \zInternalLostFlag  & -- \\
    \bottomrule
  \end{tabular}
\end{table}

\section{Self-Training Leaves Lost Slices Unrepaired on New Patients}
\label{sec:selftraining}

The self-training targets already contain the miss. On BTCV, \btcvLostPassN of them were lost
slices with the spleen labelled background (Sec.~\ref{sec:roundtrip}). Within a noise stream,
the source and self-trained models share the noise, so on a given slice the two differ only in
their weights.

\paragraph{Lost slices stay lost.}
Of the \rtLostN slices lost by the source model on the \mcohort, self-training recovered
\gstRecoveredN and left \gstStillLostN lost (\gstPersist). It also lost \gstNewLostN further
slices, raising the total to \gstLostTotalN (+\gstLostIncreasePct). Both models were run again
on a second shared noise stream with a re-implemented inference path. Of the \pairLostN slices
the source model lost on this stream, self-training recovered \pairRecoveredN, left \pairStillN
lost (\pairPersist, 95\% CI \pairPersistCI), and lost \pairNewN further slices. A shared
stream raises persistence by design. The direction of change is not inflated by it, and on both
streams self-training lost more slices than it recovered. For comparison, the source model was
re-run with new noise, which also changes the inference path. This re-run recovered
\noiseRecoveredN lost slices, left \noiseStillN lost (\noisePersist), and lost \noiseNewN
further slices. Against this baseline, self-training lost \gstNewMore more slices per patient
(95\% CI \gstNewMoreCI) and recovered \gstRecFewer fewer (\gstRecFewerCI).

On the \ccohort (\stageBN patients), the three self-trained seeds left \seedZeroPersist,
\seedOnePersist and \seedTwoPersist of the source model's lost slices lost, against
\noisePersist for the noise re-run; only the first seed shares the source noise stream. Lost
slices rose from \ccSourceLostN to \ccSeedLostRange.

\paragraph{The volumetric summary barely moves.}
Four-organ mean volumetric Dice changed by \organsMeanDiff, while lost spleen slices rose by \gstLostIncreasePct and spleen volumetric
Dice fell by only \gstSpleenDiff (95\% CI \gstSpleenDiffCI), and by \seedZeroSpleenDiff,
\seedOneSpleenDiff and \seedTwoSpleenDiff for the three seeds on the \ccohort. Severe patients rose
from \amosSevereN to \gstSevereN, not significantly (paired exact test, $p=\gstSevereTableP$).

\begin{figure}[t]
  \centering
  \includegraphics[width=\textwidth]{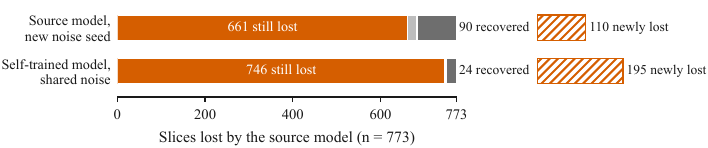}
  \caption{Self-training recovers fewer lost slices than re-drawing the noise, and loses more.
    Fate of the \rtLostN spleen slices lost by the source model on the \mcohort.
    Light grey marks partial recovery, 10--50\% coverage (\noisePartialN and \gstPartialN).
    Hatched bars, on the same scale, count slices that were not lost before.}
  \label{fig:fate}
\end{figure}

\section{Chain-and-Fill, Baselines, and Second Models}
\label{sec:repair}

\paragraph{Method.}
\cnf needs no model. It keeps 2D spleen components of at least \areaThr pixels as anchors,
links anchors whose slices are at most four apart into chains, and keeps the chain with the
most voxels. It then fills missing slices from their neighbours. Each candidate is the
neighbouring slice's mask dilated or eroded by up to 16 pixels. A candidate is accepted only if
its mean intensity lies within one standard deviation of the chain's spleen intensity, its
intensity spread stays below 1.5 times that deviation, fewer than half of its pixels belong to
other organs, and the filled slice stays within 1.2 times the source area. Filling repeats
for at most 40 rounds. \Chainonly, the same chain without filling, isolates the fill step. The
chaining rule was written after inspecting anchor chains on BTCV development patients, and the
variant was then selected on the \mcohort and frozen.

\paragraph{Confirmation.}
On the \ccohort, applied to the outputs of three self-trained seeds, \cnf met all four
pre-registered endpoints (Table~\ref{tab:repair}), including a median severe-group gain of
\cnfCone; the precision threshold of 0.50 was set after development results. In all,
\lfourRecoveredN of \lfourLostN lost slices were recovered (\lfourRecoveredPct). \Chainonly gained \chainonlyCone, so the gain comes from
filling. In severe-group gain, \cnf exceeded \nsi by \lfourVsLone (\lfourVsLoneCI).

\paragraph{Baselines.}
SAM\,2 mask propagation~\cite{ravi2024sam2} and shape interpolation start from the same
\chainonly output, and the multi-sample union combines the self-trained model's prediction with
four further noise samples. All use official defaults or settings fixed before the run, and
none was tuned. The comparison was registered after the \cnf result was known and run on the
same patients. In severe-group gain, \cnf exceeded shape interpolation by \dInterp (95\% CI
\dInterpCI) and the union by \dUnion (\dUnionCI), and could not be distinguished from SAM\,2
(\dSamTwo, \dSamTwoCI). SAM\,2 recovered more lost slices than \cnf, \samtwoRecInner against
\cnfRecInner between anchors and \samtwoRecOuter against \cnfRecOuter beyond the first or last
anchor, but only \samtwoCthree of the voxels it added were correct, and it harmed \samtwoHarmed
patient--seed pairs against \cnfHarmed.

\paragraph{Second models.}
No independent model reached usable accuracy on AMOS, so the rate of misses outside SymmAdapt
remains open. An nnU-Net~\cite{isensee2021nnunet} trained on the same MRI without adaptation
reached a spleen Dice of \mtwoGateDice on held-out MRI but \mtwoSpleenDice on CT. We then adapted two source-free discriminative models to
the same BTCV images: DFG~\cite{huai2025dfg} and, as the single pre-registered fallback,
ProtoContra~\cite{yu2023protocontra}. The \mcohort was to be opened only if spleen and
four-organ Dice reached \targetGateDice on five AMOS patients outside all cohorts. Both models
stopped at this check, with spleen Dice \dfgSpleenDice and \pcSpleenDice and four-organ Dice
\dfgFourDice and \pcFourDice. DFG takes its pseudo-labels from MedSAM, whose training data
include AMOS CT, so this overlap could only have favoured it.

\begin{table}[t]
  \centering
  \caption{Repairs on the \ccohort (\stageBN patients, three self-trained seeds; means over
    seeds, 95\% CIs from patient-level bootstrap). Gated interpolation: \nsi. Harmed: patient--seed pairs losing
    more than 0.05 Dice. Met: pre-registered endpoints met (severe gain, non-severe change, precision,
    whole-cohort gain).}
  \label{tab:repair}
  \scriptsize
  \setlength{\tabcolsep}{3pt}
  \begin{tabular}{@{}lcccccc@{}}
    \toprule
    Method & \makecell{Severe gain\\(median)} & \makecell{Non-severe\\change} &
      \makecell{Added-voxel\\precision} & \makecell{Whole-cohort\\gain} & Harmed & Met \\
    \midrule
    \cnf            & \pci{\cnfCone}{\cnfConeCI} & \cnfCtwo & \pci{\cnfCthree}{\cnfCthreeCI} & \pci{\cnfCfour}{\cnfCfourCI} & \cnfHarmed & \cnfMet \\
    \Chainonly      & \pci{\chainonlyCone}{\chainonlyConeCI} & \chainonlyCtwo & -- & \pci{\chainonlyCfour}{\chainonlyCfourCI} & \chainonlyHarmed & \chainonlyMet \\
    Gated interpolation & \pci{\nsiCone}{\nsiConeCI} & \nsiCtwo & \pci{\nsiCthree}{\nsiCthreeCI} & \pci{\nsiCfour}{\nsiCfourCI} & \nsiHarmed & \nsiMet \\
    SAM\,2 propagation  & \pci{\samtwoCone}{\samtwoConeCI} & \samtwoCtwo & \pci{\samtwoCthree}{\samtwoCthreeCI} & \pci{\samtwoCfour}{\samtwoCfourCI} & \samtwoHarmed & \samtwoMet \\
    Shape interpolation & \pci{\interpCone}{\interpConeCI} & \interpCtwo & \pci{\interpCthree}{\interpCthreeCI} & \pci{\interpCfour}{\interpCfourCI} & \interpHarmed & \interpMet \\
    Multi-sample union  & \pci{\unionCone}{\unionConeCI} & \unionCtwo & \pci{\unionCthree}{\unionCthreeCI} & \pci{\unionCfour}{\unionCfourCI} & \unionHarmed & \unionMet \\
    \bottomrule
  \end{tabular}
\end{table}

\section{Discussion and Conclusion}
\label{sec:discussion}

\paragraph{What the audit implies.}
Both gates fail for structural reasons that do not depend on the cohort. The \offgate takes
the pseudo-label it checks as its reference (\Pone), and volumetric Dice depends only on how
much volume is missed, not where (\Ptwo). Because any agreement-based gate is blind to a
reproduced miss, closing the gap requires evidence from outside the round trip, such as an
anatomical prior or a cross-slice rule like \cnf. Pseudo-label filters that keep a pixel where
predictions agree, across dropout samples~\cite{chen2021dpl}, perturbed prediction
heads~\cite{wu2023uplsfda}, or an image and its style-translated copy~\cite{wu2024fplplus},
share this property by construction and keep a miss common to all compared predictions as
reliable background. The real image alone, as nnQC uses it, did
not suffice here. We therefore recommend reporting the number of lost slices, or a presence rate,
alongside volumetric Dice, and scoring an organ when it is present in either mask, which costs
nothing. Round-trip agreement is not evidence that a pseudo-label is complete.

\paragraph{Limitations.}
We study one model family, the slice-wise flow-matching segmenter used by SymmAdapt. No
independent model reached usable accuracy on AMOS, so we have neither a test of generality nor a
reference rate of misses for a working segmenter. TotalSegmentator~\cite{wasserthal2023totalseg}
was not run because its training data may include AMOS, and an MRI-only nnQC exceeded its
pre-set training budget. Seeds are treated as fixed, slice-level fate is tracked for one seed on
two noise streams, and the noise baseline changes noise and inference path together. The spleen
is the primary organ. Nothing here shows whether the miss is specific to generative models.

\paragraph{Conclusion.}
In SymmAdapt, whole-slice misses pass both quality gates and survive self-training. Counting
lost slices exposes them, and a model-free repair recovers part.

\bibliographystyle{splncs04}
\bibliography{refs}

\end{document}